\def\year{2022}\relax
\documentclass[letterpaper]{article} 
\usepackage{aaai22}  
\usepackage{times}  
\usepackage{helvet}  
\usepackage{courier}  
\usepackage[hyphens]{url}  
\usepackage{graphicx} 
\usepackage{natbib}  
\usepackage{caption} 
\DeclareCaptionStyle{ruled}{labelfont=normalfont,labelsep=colon,strut=off} 
\usepackage{algorithm}
\usepackage{algorithmic}

\usepackage{bm}

\usepackage{booktabs}

\usepackage{newfloat}
\usepackage{listings}
\floatstyle{ruled}
\newfloat{listing}{tb}{lst}{}
\floatname{listing}{Listing}

\usepackage{xcolor}
\definecolor{neonfuchsia}{rgb}{1.0, 0.25, 0.39}

\usepackage[hang,flushmargin]{footmisc}
\newcommand\workshopnote[1]{\renewcommand\thefootnote{}\footnote{#1}}

\title{Commonsense-Aware Prompting for Controllable Empathetic Dialogue Generation}

\author {
    Yiren Liu \textsuperscript{\rm 1},
    Halil Kilicoglu \textsuperscript{\rm 2}
}
\affiliations {
    \textsuperscript{\rm 1} Informatics, University of Illinois Urbana-Champaign\\
    \textsuperscript{\rm 2} School of Information Sciences, University of Illinois Urbana-Champaign\\
    yirenl2@illinois.edu, halil@illinois.edu
}

\usepackage{bibentry}

\begin{document}

\maketitle

\begin{abstract}
Improving the emotional awareness of pre-trained language models is an emerging important problem for dialogue generation tasks. 
Although prior studies have introduced methods to improve empathetic dialogue generation, few have discussed how to incorporate commonsense knowledge into pre-trained language models for controllable dialogue generation. 
In this study, we propose a novel framework that improves empathetic dialogue generation using pre-trained language models by 1) incorporating commonsense knowledge through prompt verbalization, and 2) controlling dialogue generation using a strategy-driven future discriminator. 
We conducted experiments to reveal that both the incorporation of social commonsense knowledge and enforcement of control over generation help to improve generation performance. Finally, we discuss the implications of our study for future research.
\end{abstract}

\workshopnote{Accepted to Workshop on Knowledge Augmented Methods for Natural Language Processing, in conjunction with AAAI 2023.}

\section{Introduction}
Empathetic dialogue generation has been an important task found to be beneficial for the applications of conversational agents in many domains, such as healthcare \cite{kim_effects_2004, nadarzynski_acceptability_2019, athota_chatbot_2020} and mental health consulting \cite{lee_i_2020}. 
Recent research about pre-trained language models (PLMs, e.g., T5 and GPT-2) has advanced task performance related to dialogue generation. However, such language models still struggle to comprehend commonsense knowledge and emotions within dialogue contexts, as shown in Table. \ref{Exp:Empathy}.
        
Research \cite{li-etal-2022-kemp} has pointed out that the ability to understand users’ emotions is crucial for dialogue systems to perform empathetic conversations. For humans, the perception of emotions is based on understanding and inference of commonsense knowledge that is often implicit during conversational interactions \cite{brownLanguageModelsAre2020}. 
However, more studies are needed to understand how commonsense knowledge can be incorporated into pre-trained language models during dialogue generation tasks. 
A promising avenue in this respect is prompt-based methods, which enable direct injection of external knowledge by augmenting model input. In this work, we conduct experiments to provide further insight into whether and how commonsense knowledge injection using prompt-based methods can benefit the language model’s emotional awareness.

\begin{table}[]
\resizebox{\columnwidth}{!}{
\begin{tabular}{ll}
\hline
\textbf{Speaker:} &
  \textit{\begin{tabular}[t]{@{}l@{}}
  It was 100\% their fault \\ but they hit the water barrels and survived. \\ They had no injuries \\ but they almost ran me off the road.\end{tabular}} \\
\textbf{Listener:} &
  Did you suffer any injuries? \\
\textbf{Speaker:} &
  \textit{\begin{tabular}[t]{@{}l@{}}No I was not hit. \\ It turned out they were drunk. \\ I felt guilty \\ but realized it was his fault.\end{tabular}} \\
\textbf{\begin{tabular}[c]{@{}l@{}}Listener\\ (GPT-J 6B):\end{tabular}} &
  \textbf{Why was it your fault?} \\ \hline
\multicolumn{2}{l}{\begin{tabular}[t]{@{}l@{}}\textbf{Prompt:}\\ The speaker was almost run over by a group with a vehicle. \\ The speaker reacts with guilt.\end{tabular}} \\
\multicolumn{2}{c}{......} \\
\textbf{\begin{tabular}[c]{@{}l@{}}Listener\\ (GPT-J 6B):\end{tabular}} &
  \textbf{Are you still angry at them?} \\ \hline
\end{tabular}
}
\caption{Example of how prompting commonsense knowledge can improve the empathy of dialogue generation.}
\label{Exp:Empathy}
\vspace{-0.5cm}
\end{table}

Additionally, Plug-and-Play methods \cite{Dathathri2020Plug} are also discovered recently as a promising direction for text generation. \citet{yang_fudge_2021} introduced the method of using future discriminators (FUDGE), which enables conditioning of text generation over a given attribute based on pre-trained language models. The method has been evaluated over several different text generation tasks, including poetry generation, topic-based generation and formality transfer \cite{yang_fudge_2021}. However, few have discussed the possibility of using plug-and-play methods to control generation in a dialogue context.

Two main contributions of this study include:

\begin{itemize}
\item introducing a prompt-based method to incorporate social commonsense knowledge into pre-trained language model for dialogue generation;
\item proposing a strategy-controlled dialogue generation method that can be used on language models with or without finetuning.
\end{itemize}

\section{Related Work}
\subsection{Using Commonsense Knowledge for Empathetic Dialogue Generation}
Recent works have explored the ability of pre-trained language models in dialogue generations \cite{zhang_dialogpt_2020}.
Prior works have shown that these models tend to have difficulty in comprehending emotional expressions from users \cite{liAdversarialLearningNeural2017}. \citet{sabourCEMCommonsenseawareEmpathetic2021} considered both commonsense knowledge (cognitive) and emotional factors (affective) when encoding knowledge for empathetic dialogue generation using a seq2seq model. 
This work is built upon the cognitive and affective aspects of the theory of empathy from \cite{davisMeasuringIndividualDifferences1983}.
Later work \cite{li-etal-2022-kemp} also explored a graph-based approach for encoding commonsense knowledge in combination with emotional intensity based on VAD (Valence, Arousal and Dominance) vectors.
However, these works did not use PLMs but traditional encoder-decoder models that need to be trained from scratch, thus limiting their comprehension ability to the training dataset used. 


\subsection{Controllable Dialogue Generation}
Prior studies have also discussed using external signals to condition dialogue generation. 
\citet{xu_towards_2018} proposed the idea of controlling dialogue generation using dialogue acts, which is found to significantly improve the quality of generated dialogues. 
\citet{wang_topic_2020} has also explored the possibility of using topics to guide dialogue generation.
Further work by \citet{wu_controllable_2021} focused on using external knowledge from Wikipedia to avoid fact hallucination in dialogue generation. 
\citet{yang_fudge_2021} proposed the method of FUDGE, an attribute future discriminator, that enables strong plug-and-play control over text generation.
In this work, we propose to use plug-and-play methods in order to control empathetic dialogue generation, which enables controllable generation with or without finetuning the PLM. 


\section{Preliminaries}
\subsection{Problem Formulation}
\subsubsection{Empathetic Response Generation}
We formulate the task of empathetic dialogue response generation as follows: Given the dialogue history $C = [X_1, X_2, ..., X_{T-1}]$, where $T$ is the total number of dialogue utterances and $X_i$ is the $i$-th utterance text sequence, generate the next dialogue response $Y$. Each utterance $X_i = [x_1, x_2, ..., x_m]$ and the target utterance $Y = X_{T} = [y_1, y_2, ..., y_n]$, where $m$ and $n$ denote the total number of word tokens of each utterance.

\subsubsection{Dialogue Strategy Prediction}
The generation of empathetic responses can benefit from the guidance of proper dialogue strategies. In this work, we refer to the conversational skills introduced in \citet{hill_helping_2009} needed in delivering emotional support as dialogue strategies \cite{xu_towards_2018}, e.g. \textit{providing suggestions} and \textit{self-disclosure}. 
We denote the total dialogue strategy set as $S = \{s_i | 1 \leq i \leq T\}$, where $T$ is the total number of strategies available. With the dialogue history $C$, we need to predict the next correct dialogue strategy $s_Y \in S$ as a conditional signal for generating the correct response $Y$.

\subsection{COMET Commonsense}
We utilize the COMET-ATOMIC$_{20}^{20}$ \cite{hwangCometAtomic20202021} commonsense reasoning language model to augment dialogue history input with social-related commonsense knowledge. 
The commonsense knowledge is represented as relation-entailment tuples. For example, $([xReact, depressed])$ indicates that the subject in the dialogue (PersonX) has the reaction of feeling depressed. 
The COMET-ATOMIC$_{20}^{20}$ dataset contains 23 relation types that are categorized into three categorical types including social commonsense relations, physical relations, and event-centric relations.
In this study, we use dialogue history as the context, and used the COMET-BART model pre-trained on the COMET-ATOMIC$_{20}^{20}$ dataset to generate the commonsense entailment.

\section{Approach}
\begin{figure*}[h!]
  \centering
  \includegraphics[width=0.8\linewidth]{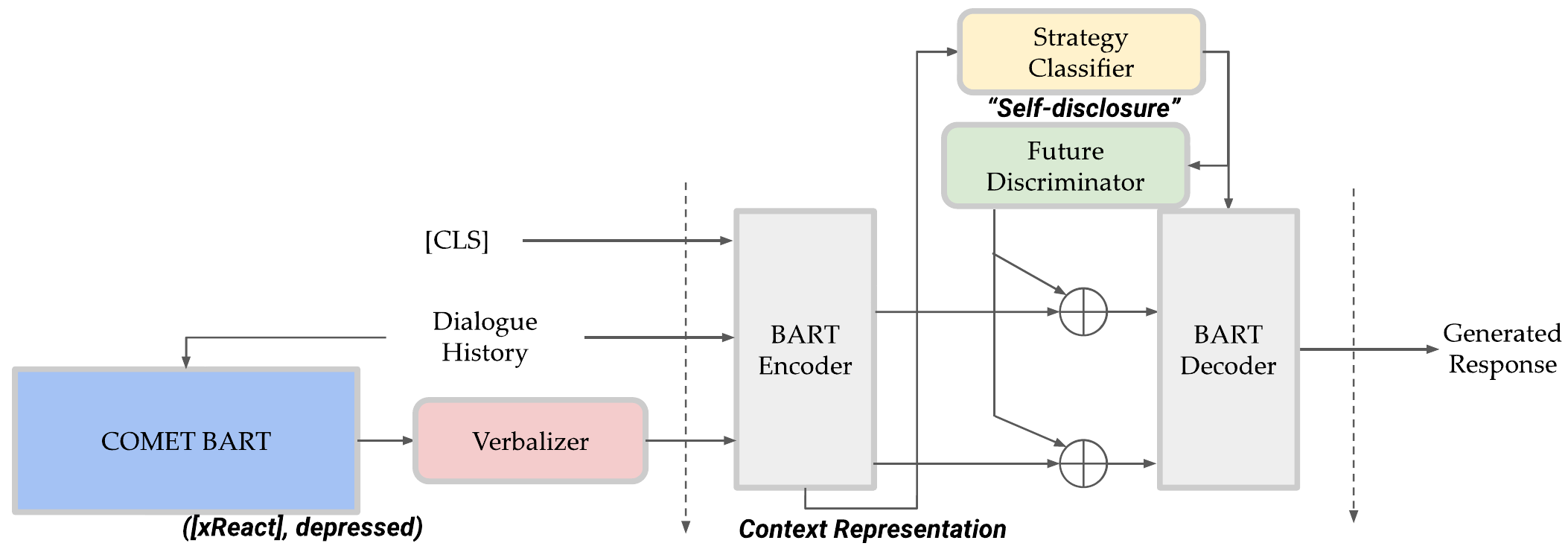}
  \caption{The Overall Framework of the Proposed Model.}
  \label{fig:Framework}
  \vspace{-0.5cm}
\end{figure*}

As shown in Figure. \ref{fig:Framework}, 
Our proposed method consists of three major components: 1) commonsense prompting, 2) dialogue strategy predictor and 3) FUDGE-controlled response generation. 

\subsection{Commonsense Prompt Verbalizer}
We first use the BART model pre-trained on the COMET-ATOMIC$_{20}^{20}$ dataset to obtain implicit commonsense knowledge that can be deduced from the dialogue context. 
We decode the commonsense entailment using 10 social relations from COMET-ATOMIC$_{20}^{20}$, including $[oEffect]$, $[oReact]$, $[oWant]$, $[xAttr]$, $[xEffect]$, $[xIntent]$, $[xNeed]$, $[xReact]$, $[xReason]$ and $[xWant]$. Relations starting with $x$ refers to the main subject of the conversation, and $o$ refers to others besides the subject. The commonsense knowledge is represented in the form of relation-entailment tuples $(r_{i,j}, e_{i,j})$ for each dialogue history $X_i \in C$, where $1 \leq j \leq 10$ refers to the $j$-th social relation. We generate entailment with all 10 relations for each dialogue history. 

To better help the language model to comprehend the commonsense relations obtained from COMET-BART, we verbalize the commonsense tuples with natural language templates using the templates proposed by \citet{hosseini_knowledge-augmented_2022}. For example, we convert the tuple $([xReact], depressed)$ into a sentence \textit{``As a result, PersonX feels depressed.''}. We verbalize each tuple obtained from entailment generation, and append it to dialogue history as the input text.

\subsection{Dialogue Strategy Predictor}
To provide stronger control over the response generation, we first need to identify the correct strategy to be deployed in the next utterance given an existing dialogue history. 
We approach the problem as a text classification task. In addition to using the LM for generating the strategy with constrained decoding, we propose two methods to predict the next utterance strategies. First, we use the complete dialogue histories as the input text sequences and ground truth strategies as the labels to train an end-to-end text classification model. This method is agnostic to the generation LM and can ideally be swapped with any off-the-shelf text classification models available. 
The other approach is to adopt the encoder of the generation LM and use the encoder representation of the dialogue history text to train a classification module jointly with a generation objective. In this case, the training loss of the generation model for each utterance can be written as a combination of classification and generation loss as follows:

\begin{equation}
    \ell_{strategy} = - log\ p(s |Encoder([x_1, x_2, ..., x_{t-1}]))
\end{equation}
\vspace{-0.4cm}
\begin{equation}
    \ell_{LM} = - \sum_{t=1}^{n} log\ p(x_t|x_1, x_2, ..., x_{t-1}, s)
\end{equation}
\vspace{-0.1cm}
\begin{equation}
    \ell = \ell_{LM} + \alpha \ell_{strategy}
\end{equation}

where $x_i$ is the $i$-th token within an utterance, $\alpha$ is the weight used to control the emphasis of strategy prediction over text generation. Note that strategy $s$ remains the same for each target utterance.
In this study, we experimented using a simple dense layer module as the strategy predictor, in which case we extract the encoder hidden representation at the [CLS] position as the context vector. 

\subsection{Future Discriminator}
We then apply additional mechanisms to control the response generation with the predicted strategy. We adopt the method of FUDGE \cite{yang_fudge_2021} to enforce the strategy control. We introduce a future discriminator specifically for identifying dialogue strategies. 

The general goal of the future discriminator is to classify the potential strategy category of a given dialogue utterance at each step of the token sequence. More specifically, for an utterance text sequence $X_i = [x_1, x_2, ..., x_n]$, we train a text classifier to predict the ground truth strategy type of $\{[x_1, x_2, ..., x_t] | 1 \leq t \leq n\}$ at every token step. Namely, we optimize for the log-likelihood sum for sequences from the first word to each word in each utterance with the loss:

\begin{equation}
    \ell_F = - \sum_{t=1}^n log\ p(s|x_1, x_2, ... , x_t)
\end{equation}

where $n$ is the length of the target utterance sequence. While the objective is determined, the model used for classification can be any model separately trainable for text classification tasks. In our experiments, we follow \citet{yang_fudge_2021}'s approach and discuss the performance of controlled generation using a light-weight LSTM text classifier, which requires a comparatively low computational cost to train and has been found by prior work \cite{schuurmans_intent_2020} to have strong performance over dialogue intent classification task.  

\begin{table*}[!ht]
\centering
\resizebox{0.85\textwidth}{!}{
\begin{tabular}{@{}cccccccc@{}}
\toprule
\textbf{Model} &
  \textbf{BLEU-1} &
  \textbf{BLEU-2} &
  \textbf{BLEU-3} &
  \textbf{BLEU-4} &
  \textbf{ROUGE-L} &
  \textbf{BERTScore} &
  \textbf{Strategy Accuracy} \\ \midrule
\textbf{BART}                  & 17.79          & 5.74          & 2.43          & 1.18          & 14.15          & 90.77          & 30.97\% \\
\textbf{COMET-BART}            & 16.39          & 6.73          & 3.47          & 2.11          & 16.23          & 90.97          & 31.94\% \\ \midrule
\multicolumn{8}{c}{\textbf{With COMET input (verbalized with templates)}}                                                                   \\ \midrule
\textbf{BART}                  & 16.31          & 5.90          & 2.71          & 1.45          & 15.31          & 90.86          & 30.46\% \\
\textbf{COMET-BART} &
  \textbf{18.53} &
  \textbf{7.80} &
  \textbf{4.12} &
  \textbf{2.54} &
  \textbf{17.79} &
  \textbf{91.21} &
  30.40\% \\ \midrule
\multicolumn{8}{c}{\textbf{With strategy classifier (linear classifier with classification + LM loss)}}                                            \\ \midrule
\textbf{BART}                  & \textbf{20.12} & \textbf{8.19} & \textbf{4.16} & \textbf{2.47} & \textbf{15.67} & 90.97          & 16.29\% \\
\textbf{COMET-BART}            & 9.28           & 3.90          & 2.13          & 1.35          & 13.56          & 90.36          & 16.29\% \\
\textbf{BART (+ FUDGE)}        & 17.53          & 5.83          & 2.51          & 1.24          & 14.29          & \textbf{91.00} & 16.29\% \\
\textbf{COMET-BART  (+ FUDGE)} & 6.60           & 2.11          & 1.05          & 0.59          & 11.82          & 90.72          & 16.29\% \\ \midrule
\multicolumn{8}{c}{\textbf{With strategy classifier (BERT classifier, trained separately from the LM)}}                                     \\ \midrule
\textbf{BART (+ FUDGE)}        & 14.00          & 5.53          & 2.96          & 1.86          & 14.86          & 90.09          & 18.46\% \\
\textbf{COMET-BART  (+ FUDGE)} & 4.74           & 1.71          & 0.86          & 0.52          & 11.36          & 90.08          & 18.46\% \\ \bottomrule
\end{tabular}
}
\caption{The experimental results. The \textbf{Strategy Accuracy} column refers to the accuracy of predicted dialogue strategies. For the linear classifier, we train the classifier using a joint loss with LM, while the strategy classifier is trained separately. }
\label{tab:all_results}
\vspace{-0.2cm}
\end{table*}

\begin{table*}[!h]
\resizebox{\textwidth}{!}{
\begin{tabular}{@{}lcl@{}}
\toprule
                    & \textbf{Dialogue strategy}        & \multicolumn{1}{c}{\textbf{Generated response}}                                               \\ \midrule
\textbf{BART} & {[}Providing Suggestions{]} & I understand that. I have been in a similar situation before and it's hard to know what to do. \\
\ \ \ \ \textbf{+ FUDGE oracle}    & {[}Self-disclosure{]}             & I had a friend die in a car accident and I was able to talk to him about how to deal with it. \\ \midrule
\textbf{COMET-BART} & {[}Affirmation and Reassurance{]} & I can understand how you feel.                                                                \\
\ \ \ \ \textbf{+ FUDGE oracle}    & {[}Self-disclosure{]}             & I had a friend die in a car accident and I was there with him, and I was able to talk to him. \\ \bottomrule
\end{tabular}
}
\caption{Examples of generated Responses w/o FUDGE control.}
\label{Tab:Exp_control}
\vspace{-0.5cm}
\end{table*}

\begin{table}[!h]
\centering
\resizebox{\columnwidth}{!}{
\begin{tabular}{lcccc}
\hline
\textbf{Model}              & \textbf{B-2} & \textbf{B-4} & \textbf{R-L} & \textbf{BERTScore} \\ \hline
\textbf{BART + oracle }              & 5.02         & 1.32         & 18.34        & 90.59              \\
\textbf{COMET-BART + oracle}         & 3.83         & 0.75         & 12.04        & 90.34              \\
\textbf{BART + oracle + FUDGE }      & 7.46         & 2.01         & 21.18        & 91.11              \\
\textbf{COMET-BART + oracle + FUDGE} & 5.76         & 1.78         & 20.82        & 90.66              \\ \hline
\end{tabular}
}
\caption{Generation results using oracle strategies.}
\label{tab:oracle_results}
\vspace{-0.5cm}
\end{table}

\section{Experimental Settings}

We experiment with our proposed method using the empathetic dialogue dataset ESConv \cite{DBLP:conf/acl/LiuZDSLYJH20}. The ESConv dataset contains a total of 1,053 multi-turn dialogues with 31,410
utterances. The dataset consists of dialogue history between pairs of help-seekers and emotional support providers, including the cause and annotated dialogue strategies. 

We compare the results with BART \cite{lewis_bart_2019} text generation model without any control. We use the BART-large model pre-trained with the objective of text de-noising using corrupted input. 
In addition to BART, we also experimented with the COMET-BART model introduced by \citet{hwangCometAtomic20202021}. Since COMET-BART is a text generation model pre-trained for commonsense reasoning, by further fine-tuning the model for dialogue generation we want to examine whether COMET-BART has a stronger capability in understanding commonsense knowledge input we provided. 
We use automatic evaluation metrics including BLEU \cite{papineni_bleu_2002}, ROUGE-L \cite{lin_rouge_2004}, and BERTScore \cite{zhang_bertscore_2020} to measure the dialogue response generation performance.




\section{Results and Discussion}

\subsection{Main Results}
We investigate the extent to which the introduction of commonsense knowledge input and controlled generation can improve the generation of dialogue response. Our results are shown in Table. \ref{tab:all_results}. Detailed comparisons of generated examples are also provided in the Technical Appendix.

\textbf{\textit{Incorporating commonsense knowledge helps the generation.}}
We inform the language models with commonsense knowledge by concatenating verbalized COMET-generated reasoning knowledge with the dialogue history. 
By prompting the BART models with verbalized COMET commonsense knowledge, the results of models \textbf{with COMET input} in Table. \ref{tab:all_results} show consistently improved results in both BART and COMET-BART models. COMET-BART is observed to perform better than vanilla BART regardless of whether the commonsense knowledge is explicitly prompted. 
This might imply that 1) commonsense knowledge learned by the COMET-BART during pretraining already benefits dialogue response generation; 2) additionally, prompting commonsense knowledge further improves generation performance, but COMET-BART comprehends the knowledge better than the vanilla BART model since it has already been pre-trained on commonsense reasoning task.

\textbf{\textit{FUDGE strategy control improves the generation.}} We compare the generation results from baseline models with models controlled with FUDGE. We first provide ground truth oracle strategies to condition the generation for comparison. As shown in the results of \textbf{BART/COMET-BART + oracle + FUDGE} in Table. \ref{tab:oracle_results}, using FUDGE as a control for dialogue generation is able to improve the performance compared to simply using the language models. This implies that FUDGE enforces stronger control over generation given the correct strategy, which can also be observed from the examples shown in Table. \ref{Tab:Exp_control}. 
However, this means generation would largely rely on the strategy predictor. 
This corresponds to the results listed in Table. \ref{tab:all_results}. Using a separate strategy classifier does not fulfill the goal of correctly predicting the strategy, thus leading to worse generation results. Future work needs to be done to improve the performance of dialogue strategy prediction.

\textit{\textbf{Jointly training with generation and strategy prediction loss improves generation performance on vanilla BART.}} Another interesting observation is that when finetuning the language model with a joint loss of generation and strategy classification, the performance of vanilla BART improved when not using FUDGE control (as shown in the \textbf{BART with strategy classifier} results in Table. \ref{tab:all_results}). The strategy prediction accuracy dropped while generation performance improved. This might be because separating the training objective allows models to have stronger adaptability to generate strategy-specific responses. However, this improvement was not observed in the COMET-BART model. 


\section{Conclusion}
In this work, we introduced a new language model-based framework for controllable empathetic dialogue response generation. We use experiment results on a public dataset to show that: 1) Our method of incorporating commonsense knowledge through verbalized prompts can improve the dialogue generation quality, 2) By separating the dialogue strategy prediction with the generation, we can use future discriminator to enforce stronger control over generated dialogue responses. 
Our method can be used on off-the-shelf generative language models with or without fine-tuning.
Future work can further explore how dialogue strategy prediction can be improved to benefit dialogue generation, and introduce better prompting methods in addition to using templates to verbalize commonsense knowledge.

\bibliography{references_new}

\newpage
\appendix

\begin{table*}[]
\centering
\begin{tabular}{p{3cm}p{14cm}}
\hline
\textbf{Context} &
  \textbf{Here! Last Christmas I bought the condo of my dreams. This April 2020 I lost my job due Covid 19. My unemployment is ending and my savings are depleted. I've recived an eviction notice and don't know what to do. Help!} \\ \hline
\textbf{Seeker} &
  Have you ever been evicted? \\
\textbf{Helper} &
  {[}Others{]} Where are you that you are able to be evicted? \\
\textbf{Helper} &
  {[}Providing Suggestions{]} You should not be able to be evicted from your home during this time in pretty much every state. Have you tried contacting anyone for help other than your landlord? \\
\textbf{Seeker} &
  I'm still in my condo but I no longer have a job, umemployment has run out and my savings are all gone. \\
\textbf{Helper} &
  {[}Providing Suggestions{]} Well the plus side is they cannot kick you out right now. Have you talk to your landlord about working out some payment arrangement. Have you tried applying for any job? Now is not the time for pride to get in the way, i would accept anything \\
\textbf{Seeker} &
  When you say work out a plan what do you mean? I have no job. I have been applying for jobs but I haven't received a callback or an offer of employment. Have you ever been in this situation? \\
\textbf{Seeker} &
  Also, I'm calling a helpline so Pride is not my issue. \\
\textbf{Helper} &
  {[}Others{]} I have but the referendum on evictions with covid makes this situation different. I am sorry if that pride comment came off as crass. I was just trying to say desperate times come with desperate measures \\ \hline
\multicolumn{2}{c}{\textbf{Ground truth}} \\ \hline
\textbf{Helper} &
  {[}Providing Suggestions{]} They could work out a way you can pay extra monthly when you get back on your feet. \\ \hline
\multicolumn{2}{c}{\textbf{Baseline models}} \\ \hline
\textbf{BART} &
  {[}Providing Suggestions{]} This is not your issue, i would say that you should be proud of yourself for being able to get out of your home. I am sorry if that came off as crass. \\
\textbf{COMET-BART} &
  {[}Affirmation and Reassurance{]}  I have been in this situation. I was not able to get out of it due to covid. I was able to get out of it because I had a plan. \\ \hline
\multicolumn{2}{c}{\textbf{With COMET output (verbalized with templates)}} \\ \hline
\textbf{BART} &
  {[}Providing Suggestions{]}  i would say you should try to find a job and see if you can get a raise. I would also suggest talking to your landlord about working out some payment arrangement. \\
\textbf{COMET-BART} &
  {[}Self-disclosure{]}  I have been in your shoes. \\
\textbf{BART (+ FUDGE)} &
  {[}Providing Suggestions{]}  Have you tried contacting a local government to see if they can help you out? \\
\textbf{COMET-BART (+ FUDGE)} &
  {[}Self-disclosure{]}  my I have been in this situation too. I was in a similar situation and i was able to get out of my home and move to a new one. \\ \hline
\multicolumn{2}{c}{\textbf{With strategy classifier (Linear CLF with classification + LM loss)}} \\ \hline
\textbf{BART} &
  {[}Providing Suggestions{]} I am sorry to hear that. I have been in this situation before and it is not my issue. I would not be able to help you. I am sorry if that pride comment came off as crass. \\
\textbf{COMET-BART} &
  {[}Providing Suggestions{]}  I am sorry. I was just trying to say desperate times come with desperate measures \\
\textbf{BART (+ FUDGE)} &
  {[}Providing Suggestions{]}  I could tell you are feeling a bit down about the situation. I would suggest you try to find a job that pays more than you can afford. \\
\textbf{COMET-BART (+ FUDGE)} &
  {[}Providing Suggestions{]}  Have you tried to talk to your landlord about a payment plan? \\ \hline
\multicolumn{2}{c}{\textbf{With strategy classifier (BERT classifier, trained saparately from the LM)}} \\ \hline
\textbf{BART (+ FUDGE)} &
  {[}Question{]} I do not think you should feel ashamed of your pride. What you are doing is the right thing. \\
\textbf{COMET-BART (+ FUDGE)} &
  {[}Question{]}  Do you have any friends or family that you can talk to about this? \\ \hline
\multicolumn{2}{c}{\textbf{strategy oracle}} \\ \hline
\textbf{BART (+ FUDGE)} &
  {[}Providing Suggestions{]}  This is not your issue, i would just say that you should be prepared for the worst. I have been in this situation before and it was a lot worse than i thought. \\
\textbf{COMET-BART (+ FUDGE)} &
  {[}Providing Suggestions{]}  Have you tried to talk to your landlord about a payment plan? \\ \hline
\end{tabular}
\caption{An Example for comparison of different generation methods.}
\end{table*}

\section{Acknowledgments}
This research used the Delta advanced computing and data resource which is supported by the National Science Foundation (award OAC 2005572) and the State of Illinois. Delta is a joint effort of the University of Illinois Urbana-Champaign and its National Center for Supercomputing Applications.

\end{document}